%% file: top.tex
\documentclass[10pt,twocolumn,letterpaper]{article}

\pdfoutput=1

\usepackage[usenames,dvipsnames]{color}
\usepackage{booktabs}
\usepackage{cvpr}
\usepackage{times}
\usepackage{epsfig}
\usepackage{graphicx}
\usepackage{amssymb}
\usepackage{amsmath, amsfonts}
\usepackage{relsize}
\usepackage{subfigure}
\usepackage{algorithm}
\usepackage{multirow}
\usepackage{tabularx}
\usepackage{array}
\usepackage{caption}
\usepackage{algorithmic}
\usepackage{enumitem}
\usepackage{url}
\usepackage{mathrsfs}
\usepackage{ragged2e}
\usepackage{bbold}

\input{defs}

\usepackage[pagebackref=true,breaklinks=true,letterpaper=true,colorlinks,bookmarks=false]{hyperref}

\cvprfinalcopy 

\ifcvprfinal\fi
\begin{document}
	
\title{Flight Dynamics-based Recovery of a UAV Trajectory using Ground Cameras}

%

\author{
	$\qquad$ Artem Rozantsev\mynote{1} $\qquad$ Sudipta N. Sinha\mynote{2} $\qquad$ Debadeepta Dey\mynote{2} $\qquad$ Pascal Fua\mynote{1} \\\\
    \mynote{1}{\hspace{2pt} Computer Vision Laboratory, EPFL} $\qquad \qquad \qquad$ \mynote{2}{\hspace{2pt} Microsoft Research} \\
    {\tt\small $\quad \qquad$ \{artem.rozantsev, pascal.fua\}@epfl.ch $\quad$ \{sudipta.sinha, dedey\}@microsoft.com}\\
	}
	
\maketitle
\begin{abstract}

We propose a new method to estimate the 6-dof trajectory of a flying object such as a quadrotor UAV within a 3D airspace monitored using multiple fixed ground cameras. It is based on a new structure from motion formulation for the 3D reconstruction of a single moving point with known motion dynamics. Our main contribution is a new bundle adjustment procedure which in addition to optimizing the camera poses, regularizes the point trajectory using a prior based on motion dynamics (or specifically flight dynamics). Furthermore, we can infer the underlying control input sent to the UAV's autopilot that determined its flight trajectory.

Our method requires neither perfect single-view tracking nor appearance matching across views. For robustness, we allow the tracker to generate multiple detections per frame in each video. The true detections and the data association across videos is estimated using robust multi-view triangulation and subsequently refined during our bundle adjustment procedure. Quantitative evaluation on simulated data and experiments on real videos from indoor and outdoor scenes demonstrates the effectiveness of our method.
\end{abstract}

\input{introduction4.tex}
\input{related_work.tex}
\input{problem2.tex}
\input{dynamics_prior2.tex}

\input{optimization3.tex}
\input{experiments.tex}

\section{Conclusion}
We have presented a new technique for accurately reconstructing the 3D trajectory of a quadrotor UAV observed from multiple cameras. We have shown that using motion information significantly improves the accuracy of reconstructed trajectory of the object in the 3D environment. Furthermore our method allows inferring the internal parameters of the quadrotor, such as roll, pitch angles and thrust, that is being commanded by the operator. Inferring these parameters has a broad variety of applications, ranging from reinforcement learning to providing analytics for pilots in air race competitions and making feasible research on UAVs in large outdoor and indoor spaces without depending on expensive motion capture systems.

{\small
	\bibliographystyle{IEEEtran}
	\bibliography{string,ref}
}

\newpage
\input{appendix.tex}

\end{document}

%% file: defs.tex
\usepackage{color}
\usepackage[%
linewidth=1pt,
middlelinecolor= black,
middlelinewidth=0.4pt,
roundcorner=1pt,
topline = false,
bottomline = false,
leftmargin=0pt,
rightmargin=0pt,
skipabove=0pt,
skipbelow=0pt,
innerleftmargin=0pt,
innerrightmargin=0pt,
innertopmargin=3pt,
innerbottommargin=3pt,
]{mdframed}

\newcommand{\comment}[1]{}

\newcommand{\bX}[0]{\mathbf{X}}

\newcommand{\bC}[0]{\mathbf{C}}
\newcommand{\bc}[0]{\mathbf{c}}
\newcommand{\bP}[0]{\mathbf{O}}
\newcommand{\bp}[0]{\mathbf{o}}
\newcommand{\bE}[0]{E}

\newcommand{\bB}[0]{\mathbf{B}}

\newcommand{\bb}[0]{\mathbf{b}}

\newcommand{\bV}[0]{\mathbf{V}}
\newcommand{\bU}[0]{\mathbf{U}}
\newcommand{\bTheta}[0]{\mathbf{\Theta}}
\newcommand{\bPhi}[0]{\mathbf{\Phi}}
\newcommand{\bPsi}[0]{\mathbf{\Psi}}
\newcommand{\bGamma}[0]{\mathbf{\Gamma}}
\newcommand{\bgamma}[0]{\mathbf{\gamma}}
\newcommand{\bOmega}[0]{\mathbf{\Omega}}

\newcommand{\mycode}[1]{\texttt{#1}}

\newcommand{\bx}[0]{\mathbf{x}}

\newcommand{\bv}[0]{\mathbf{v}}
\newcommand{\ba}[0]{\mathbf{a}}

\newcommand{\Real}[0]{\mathbb{R}}

\newcommand{\fig}[1]{Fig.~\ref{fig:#1}}
\newcommand{\sect}[1]{Section~\ref{sec:#1}}
\newcommand{\figs}[2]{Figures~\ref{fig:#1} and~\ref{fig:#2}}
\newcommand{\tbl}[1]{Table~\ref{tbl:#1}}
\newcommand{\eqt}[1]{Eq.~\ref{eq:#1}}
\newcommand{\eqs}[2]{Eqs.~\ref{eq:#1} and~\ref{eq:#2}}
\newcommand{\alg}[1]{Algorithm~\ref{alg:#1}}
\newcolumntype{M}[1]{>{\centering\arraybackslash}m{#1}}
\newcolumntype{R}[1]{>{\RaggedLeft\arraybackslash}p{#1}}
\newcolumntype{L}[1]{>{\RaggedRight\arraybackslash}p{#1}}

\newcommand{\hl}[1]{{\bf #1}}

\newcommand{\mynote}[1]{\renewcommand{\thefootnote}{\alph{footnote}}\footnotemark[#1]}

%% file: introduction4.tex

\section{Introduction}

Rapid adoption of unmanned aerial vehicles (UAV) and drones for civilian applications will create demand for low-cost aerial drone surveillance technology in the near future.
Although, acoustics~\cite{Busset2015}, radar~\cite{Barton2004} and radio frequency (RF) detection~\cite{Nguyen2016} have shown promise, they are expensive and often ineffective at detecting small, autonomous UAVs~\cite{Boniger2016}. Motion capture systems such as Vicon~\cite{Vicon} and OptiTrack~\cite{Optitrack} work for moderate sized scenes. However, the use of active sensing and special markers on the target makes them ineffective for tracking non-cooperative drones in large and bright outdoor scenes. With the exception of some recent works~\cite{Rozantsev2015,Martínez2011}, visual detection and tracking of drones using passive video cameras remains a largely unexplored topic.

\begin{figure}[t!]
\centering
\includegraphics[height=1.8cm,width=0.48\linewidth]{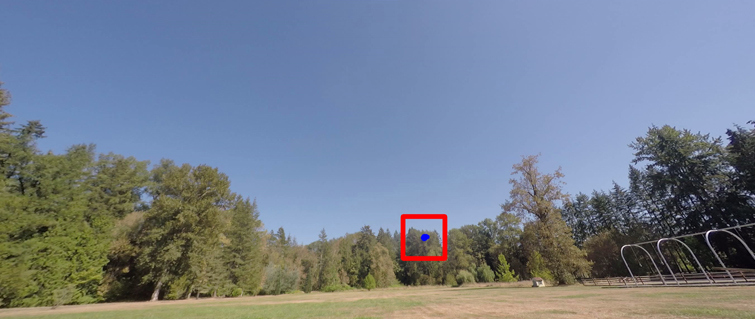}
\hspace{1mm}
\includegraphics[height=1.8cm,width=0.48\linewidth]{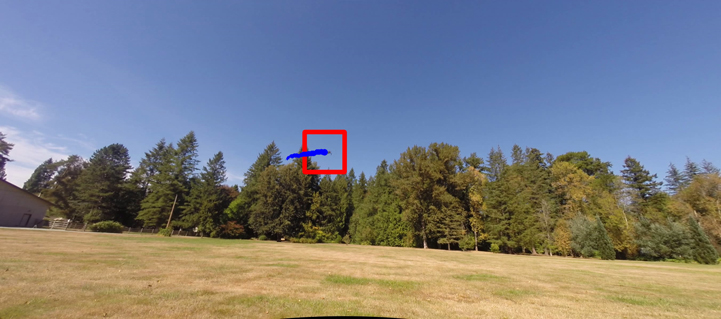}\\
\includegraphics[width=0.8\linewidth]{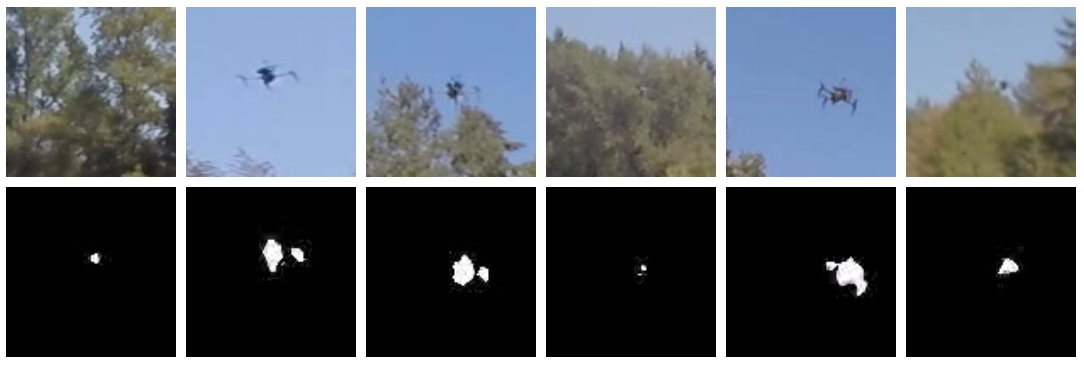}\\
\includegraphics[width=\linewidth]{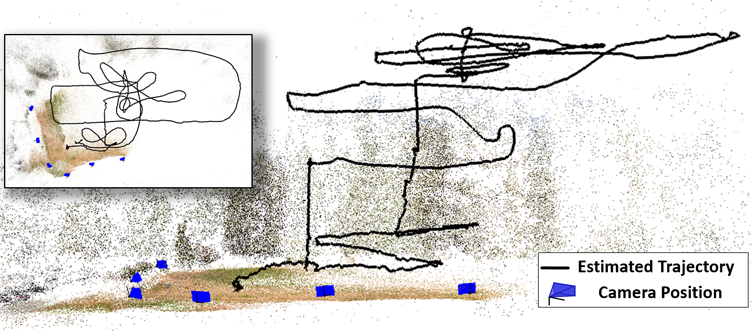}\\
\vspace{1mm}
\includegraphics[width=\linewidth]{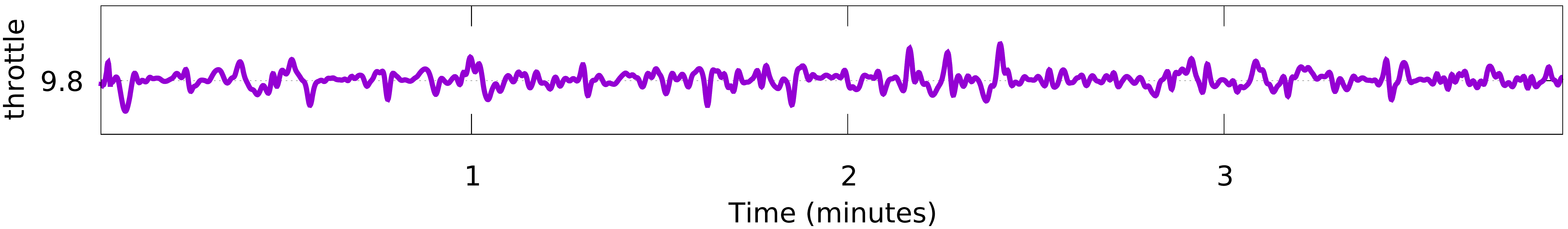}\\
\vspace{-0.1cm}
\caption{A quadrotor UAV was manually flown to a height of 45 meters above a farm within a $100 \times 50 m^2$ area with six cameras on the ground. [\textsc{Top}] Two input frames along with the detections and zoomed-in views of the UAV are shown. [\textsc{Middle}] 3D trajectory for a 4 minute flight and camera locations estimated by our method. The inset figure shows the top-down view. [\textsc{Bottom}] Estimated throttle signal (one of the control inputs sent to the autopilot).}
\label{fig:teaser}
\vspace{-0.5cm}
\end{figure}

	

Existing single-camera detection and tracking methods are mostly unsuitable for drone surveillance due to their limited field of view and the fact that it is difficult to accurately estimate the distance of objects far from the camera that occupy very few pixels in the video frame. Using multiple overlapping cameras can address these limitations.
However, existing multi-camera tracking methods are designed to track people, vehicles to address indoor and outdoor surveillance tasks, where the targets are often on the ground. In contrast, small drones must be tracked within a 3D volume that is orders of magnitude larger. As a result its image may occupy less then $20$ sq. pixels in a 4K UHD resolution video stream. Most existing multi-camera systems also rely on accurate camera calibration that requires someone to collect calibration data by walking around in the scene. A drone detection system is difficult to calibrate in this way because the effective 3D working volume is large and extends high above the ground.

In this paper, we present a new structure from motion (SfM) formulation to recover the 6-dof motion trajectory of a quadrotor observed by multiple fixed cameras as shown in \fig{teaser}. We model the drone as a single moving point and assume that its underlying flight dynamics model is known. Our contributions are three fold:
\vspace{-0.1cm}
\begin{itemize}[leftmargin=*]
	\setlength\itemsep{0.0em}
	\setlength{\parskip}{2pt}
	\item We propose a novel bundle adjustment (BA) procedure that not only optimizes the camera poses and the 3D trajectory, but also regularizes the trajectory using a prior based on the known flight dynamics model.
	\item This method lets us explicitly infer the underlying control inputs sent to the UAV's autopilot that determined its trajectory. This could provide analytics to drone pilots or enable learning controllers from demonstration~\cite{Abbeel2010}.
	\item Finally, our BA procedure uses a new cost function. It is based on traditional image reprojection error but does not depend on explicit data association derived from image correspondences, which is typically considered a pre-requisite in classical point-based SfM.	
\end{itemize}
\vspace{-0.1cm}

\noindent
Briefly, the latter lets us keep multiple 2D detections per frame instead of a single one. The true detection is indexed using a per-frame assignment variable. These variables
are initialized using a RANSAC-based multi-view triangulation step and further optimized during our bundle adjustment procedure. This makes the estimation less reliant on either perfect single-view tracking or cross-view appearance matching both of which can often be inaccurate.

Our method runs batch optimization over all the videos, which can be viewed as a camera calibration technique that uses the drone as a calibration object. In this work, we assume
that the videos are synchronized, have known frame-rates and the cameras intrinsics and lens parameters are also known whereas an initial guess of the camera poses are available. Finally, we assume only a single drone in the scene. We evaluate our method extensively on data from a realistic quadrotor flight simulator and real videos captured in both indoor and outdoor scenes. We demonstrate that the method is robust to noise and poor initialization and consistently outperforms baseline methods in terms of accuracy. 

%% file: related_work.tex

\section{Related work}
\label{sec:related}

We are not aware of any existing method that can accurately recover a UAV's 3D trajectory from ground cameras and infer the underlying control inputs that determined its trajectory. However, we review closely related works that address single and multi-camera tracking, dynamic scene reconstruction and constrained bundle adjustment.

\vspace{0.1cm}
\noindent \textbf{Single-View Tracking.}
This topic has been well studied in computer vision~\cite{Luo15}.
However, most trackers struggle with tiny objects such as flying birds~\cite{Huang2016} and tracking multiple tiny targets remains very difficult even with infrared cameras~\cite{Betke2007}. Some recent works~\cite{Rozantsev2015,Dey2011} proposed practical sense-and-avoid systems for distant flying objects using passive cameras that can handle low-resolution imagery and moving cameras. However, these methods cannot recover accurate 3D UAV trajectories.

\vspace{0.1cm}
\noindent \textbf{Multi-View Tracking.}
Synchronized passive multi-camera systems are much more robust at tracking objects within a 3D scene~\cite{Collins2000,Collins2001}. Traditionally, they have been proposed for understanding human activities, analyzing sports scenes and for indoor, outdoor and traffic surveillance~\cite{Breitenstein09, Berclaz11, Andriyenko11a, Qin12, Shu12,Butt13}. These methods need calibrated cameras and often assume targets are on the ground, and exploit this fact by proposing efficient optimization techniques such as bipartite graph matching~\cite{Shu12,Breitenstein09,Qin12}, dynamic programming~\cite{Berclaz11,Andriyenko11a} and min-cost network flow~\cite{Butt13}. These methods have rarely been used to track tiny objects in large 3D volume, where the aforementioned optimization methods are impractical. Furthermore, conventional calibration methods are unsuitable in large scenes, especially when much of the scene is high above ground level.

\vspace{0.1cm}
\noindent \textbf{Multi-view 3D reconstruction.}
Synchronized multi-camera systems have also been popular for dynamic scene reconstruction. While most existing techniques require careful pre-calibration, some techniques make it possible to calibrate cameras on the fly~\cite{Avidan00,Kaminski04,Wedge06,Yuan06,Caspi06,Padua10,Valmadre12,Sinha2010,Sinha2004,Puwein2014,Vo16}. Avidan~\etal~\cite{Avidan00} proposed a method for simple linear or conical object motion which was later extended to curved and general planar trajectories~\cite{Kaminski04,Yuan06}. More recent methods have exploited other geometric constraints for joint tracking and camera calibration~\cite{Wedge06,Caspi06,Padua10,Valmadre12}. However, these methods require accurate feature tracking and matching across views and are not suitable for tiny objects. Sinha~\etal~\cite{Sinha2010} use silhouettes correspondence and Puwein~\etal~\cite{Puwein2014} used human pose estimates to calibrate cameras. They do not require cross view feature matching but only work on small scenes with human actors.

Vo~\etal~\cite{Vo16} need accurate feature tracking and matching but can handle unsynchronized and moving cameras. They reconstruct 3D trajectories on the moving targets using motion priors that favor motion with constant velocity or constant acceleration. While our motivation is similar, our physics-based motion dynamics prior is more realistic for UAVs and enables explicit recovery of underlying parameters such as the control inputs given by the pilot.

\vspace{0.1cm}
\noindent \textbf{Constrained Bundle Adjustment.}
In conventional bundle adjustment~\cite{Triggs2000}, camera parameters are optimized along with the 3D structure which is often represented as a 3D point cloud.
Sometimes, regularization is incorporated into bundle adjustment via soft geometric constraints on the 3D structure, including planarity~\cite{Bartoli2003}, 3D symmetry~\cite{Cohen2012}, bound constraints~\cite{Gong2015} and prior knowledge of 3D shape~\cite{Fua2000}. These priors can add significant overhead to the Levenberg-Marquardt nonlinear least squares optimization~\cite{Marquardt1963}. In our problem setting, the sequential nature of the dynamics-based prior introduces a dependency between all structure variables i.e. those representing sampled 3D points on the trajectory. This leads to a dense Jacobian and makes the nonlinear least square problem infeasible for long trajectories. In this paper, we propose an alternative approach that retains the sparsity structure in regular BA. Our idea is based on generating an intermediate trajectory and then adding soft constraints to the 
variables associated with 3D points during optimization. We discuss it in \sect{opt}.

%% file: problem2.tex

\section{Problem formulation}
\label{sec:problem}

Consider the bundle adjustment (BA) problem for point-based SfM~\cite{Hartley00}.
Given image observations $\bP = \{\bp_{jt}\} : \bp_{jt} \in \mathbb{R}^2$ of
$T$ 3D points in $M$ static cameras, one seeks to estimate the coordinates of
the 3D point $\bX = \{\bx_t\}:\bx_t \in \Real^3, t \in [1..T]$ and the camera poses
$\bC = \{\bc_j\}, j \in [1..M]$, where each $\bc_j = [\mathbf{R_j}|\mathbf{t_j}] \in \mathbb{R}^{3\times 4}$.
For our trajectories, let $\bx_t$ denote each 3D point on the trajectory at time $t$.
When we have unique observations of $\bx_t$ in all the cameras where it is visible (denoted by $\bp_{jt}$ for the
$j$-th camera at time $t$), the problem can be solved by minimizing an objective based on the 2D image reprojection error
\begin{equation}
\label{eq:repro_cost}
\bE_{BA}(\bC,\bX,\bP) = \sum \limits_{t=1}^{T} \sum \limits_{j \in \bOmega_t} \rho(\pi(\bc_j,\bx_t) - \bp_{jt}),
\end{equation}
\noindent
where $\bOmega_t \subseteq \bC$ is the set of cameras where the 3D point $\bx_t$ is visible at time $t$, $\pi(\bc_j,\bx_t) : \mathbb{R}^3 \rightarrow \mathbb{R}^2$ is the function that projects $\bx_t$ into camera $\bc_j$ and the function $\rho(\cdot) : \mathbb{R}^2 \rightarrow \mathbb{R}$ robustly penalizes reprojections of $\bx_t$ that deviate from $\bp_{jt}$.

Now, let us relax the assumption that unique observations of $\bx_t$ are available in the camera views where it is visible.
Instead, we will assume that multiple candidate observations are given in each camera at time $t$, amongst which at most one
is the true observation.
To handle this situation, we propose using a new objective of the following form:
\begin{equation}
\label{eq:repro_cost_min}
\bE(\bC,\bX,\bP) = \sum \limits_{t=1}^{T} \sum \limits_{j \in \bOmega_t} \min\limits_{k} \rho \left(\pi(\bc_j,\bx_t) - \bp_{jtk} \right) ,
\end{equation}
\noindent
where $\bp_{jtk}$ is the $k$-th amongst $K_{tj}$ candidate 2D observations at time $t$ in camera $j$.
This objective is motivated by the fact that many object detectors
naturally produce multiple hypotheses but accurately suppressing all the false detections in a single view can be quite difficult.

So far, we have treated $\bX$ as an independent 3D point cloud and ignored the fact that the points lie on a UAV's flight trajectory.
Since consecutive points on the trajectory can be predicted from a suitable motion dynamics model (given additional information about the user
inputs), we propose using such a regularizer
in our BA formulation for higher robustness to erroneous or noisy observations. Our objective function therefore takes the following form:
\begin{equation}
\label{eq:optimization_problem}
\mathop{\arg\min} \limits_{\bC,\bX,\bGamma} \left( \bE(\bC,\bX,\bP) + \lambda R(\bX,\bGamma) \right).
\end{equation}
\noindent
Here, the regularizer $R(\bX,\bGamma)$ favors trajectories that can be explained by {\em good} motion models.
$\bGamma : \{\bgamma_t\}$ denotes the set of latent variables $\bgamma_t$ for the motion model at time $t$ and $\lambda$ is a scalar weight.
Typically, such regularization, where structure variables $\{\bx_t\}$ depend on one another destroys the sparsity in the problem, which is key to efficiently solving large BA instances.
However, in our work, we avoid that issue by using regularizers of the following type.
\begin{equation}
\label{eq:prior_generic}
R(\bX,\bGamma) = \sum \limits_{t=1}^{T} (\bx_t - \hat{\bx}_t(\bgamma_t))^2,
\end{equation}
\noindent
where $\{\hat{\bx}_t\}$ are 3D points predicted by a motion model. As a simple example,
one could smooth the trajectory estimate from a previous iteration of BA by setting $\hat{\bx} = (g \ast \bx)$ with Gaussian kernel $g$ and $(\cdot \ast \cdot)$ the convolution operator, to favor a
smooth trajectory in the current iteration. There are no latent variables for this simple case and so $\bGamma = \emptyset$. Next, we discuss a more realistic case, involving a flight dynamics model for a quadrotor UAV and based on it derive an appropriate regularizer $R(\bX,\bGamma)$. 

%% file: dynamics_prior2.tex

\subsection{Flight dynamics model}

\begin{table}[h!]
	\centering
	\scalebox{0.9}{
	\begin{tabular}{ll|ll}
		\toprule
		$\bV : \{\bv_t\}$ 	& : velocity  & $\bPhi : \{\phi_t\}$	& : roll \\
		$m$ 				& : mass & 	$\bTheta : \{\theta_t\}$	& : pitch \\
		$\bU : \{u_t\}$		& : throttle & 	$\bPsi : \{\psi_t\}$	& : yaw \\
		\midrule
		$\bB : \{\bb_t\} $ & : angular velocity & $\bU_{\phi} : \{u_{\phi}(t)\}$ & \multirow{3}{*}{\begin{tabular}{l} \hspace{-0.2cm}: control \\ inputs \\ \end{tabular}}\\
		$I_x,I_y,I_z$ & : moments of inertia & $\bU_{\theta} : \{u_{\theta}(t)\}$ & \\
		$J_{tp}$ & : propeller's inertia & $\bU_{\psi} : \{u_{\psi}(t)\}$ & \\
		\bottomrule
	\end{tabular}}
	\caption{Notations for the physics-based model~\cite{Webb12}.}
	\label{tbl:notations}
\end{table}

While several flight dynamics models for quadrotors are known, we use the one proposed by Webb~\etal~\cite{Webb12,Li11}.
\tbl{notations} presents the relevant notation. Here, we only include a subset of the equations that are required for deriving the prior
or computing the control inputs. ($\bU$, $\bPhi, \bTheta$) denote the thrust and the angles for the complete trajectory.
The control inputs $[\bU, \bU_\phi, \bU_\theta, \bU_\psi]$ in \tbl{notations} denote the joystick positions in the RC controller.
In our case, we need to assume that the yaw angle is zero $\bPsi = 0, \bU_{\psi} = 0$. This implies that quadrotor is always ``looking'' in
a certain direction, regardless of the motion direction. Since propeller inertia $J_{tp}$ is often very small ($\sim 10^{-4}$), we set it to zero
to reduce the model complexity without losing much accuracy.

\noindent From the basic equations of motion, we have
\begin{equation}
\label{eq:velocity_dm}
\bx_{t+1} = \bx_t + \bv_{t}dt, \quad \bv_{t+1} = \bv_t + \ba_tdt,
\end{equation}
\noindent
where $\ba_t = (a_x(t),a_y(t),a_z(t))$ is the acceleration of the quadrotor at time $t$. From the principles of rigid body dynamics, we have the following equation.
\begin{equation}
\label{eq:acceleration_dm}
\begin{bmatrix}
a_x(t)\\
a_y(t)\\
a_z(t)\\
\end{bmatrix}
=
\begin{bmatrix}
0 \\
0 \\
-g \\
\end{bmatrix}
+
\begin{bmatrix}
\sin{\theta_t}\cos{\phi_t} \\
-\sin{\phi_t} \\
\cos{\theta_t}\cos{\phi_t} \\
\end{bmatrix}
\frac{u_t}{m},
\end{equation}
\noindent
where $g$ is the standard gravitational acceleration. From \eqt{acceleration_dm} we obtain the following expression for $(\phi_t,\theta_t,u_t)$:
\begin{equation}
\label{eq:pose_prediction}
\begin{array}{rl}
u_t &= m\sqrt{a_x(t)^2 + a_y(t)^2 + (a_z(t)+g)^2}, \\
\phi_t &= \arcsin\left(-a_y(t)m / u_t\right), \\
\theta_t &= \arcsin\left((a_x(t)m / u_t) \cos{\phi_t}\right), \\
& \phi_t \in \left[-\dfrac{\pi}{2},\dfrac{\pi}{2}\right), \theta_t \in \left[-\dfrac{\pi}{2},\dfrac{\pi}{2}\right) \\
\end{array}
\end{equation}
\noindent
Here, $u_t$ must be greater than zero which is always satisfied by a quadrotor in flight.
Finally, we can estimate the UAV's angular velocity $\bb_t = [\bb_p(t),\bb_q(t),\bb_r(t)]$ as follows:
\begin{equation}
\label{eq:body_rates}
\begin{bmatrix}
\bb_p(t)\\
\bb_q(t)\\
\bb_r(t)\\
\end{bmatrix}
=
\begin{bmatrix}
(\phi_t-\phi_{t-1})/dt \\
((\theta_t - \theta_{t-1})/dt) (\cos{\phi_t}/\sin^2{\phi_t}) \\
- \theta_t/ \sin{\phi_t} \\
\end{bmatrix},
\end{equation}
\noindent
which can be used to compute control inputs as follows:
\begin{equation}
\label{eq:user_commands}
\begin{array}{l}
u_{\phi}(t) = I_{x} \frac{\bb_p(t)-\bb_p(t-1)}{dt} - \left( I_{y} - I_{z} \right)\bb_q(t)\bb_r(t) \\
u_{\theta}(t) = I_{y} \frac{\bb_q(t)-\bb_q(t-1)}{dt} - \left( I_{z} - I_{x} \right)\bb_p(t)\bb_r(t) \\
\end{array}.
\end{equation}
\noindent
Next, we describe the flight dynamics based regularizer.

%
%

\subsection{Flight dynamics prior}
\label{sec:quad_dynamics}

In the rest of the paper, we denote $\bGamma = [\bPhi,\bTheta,\bU]^{\tau}$. These variables will
serve as the latent variables in the flight dynamics based prior (we use the term regularizer and prior interchangeably).
The dynamics model provides us two transformations denoted $\mathcal{F}$ and $\mathcal{G}$ below.
\begin{equation}
\label{eq:transform_F}
\mathcal{G}: \bX \rightarrow \bGamma, \mathcal{F}: \bGamma \rightarrow \bX.
\end{equation}
\noindent
Equations \ref{eq:velocity_dm} and \ref{eq:pose_prediction} are used to obtain $\bGamma$ from a trajectory estimate.
Similarly, the values of $\bX$ can be derived from $\bGamma = (\bU, \bPhi, \bTheta)$ by recursively using \eqs{velocity_dm}{acceleration_dm}.
This is equivalent to performing integration with respect to time which uniquely determines the quadrotor's internal state variables (position, velocity, acceleration etc.)
up to constant unknown namely the quadrotor's state at time $t = 0$.

The general idea of the regularizer will be to add appropriate constraints to the latent variables $(\bU, \bPhi, \bTheta)$ during the intermediate steps of the bundle adjustment
procedure. Below, we use $\mathcal{H}$ to denote such a function.
\begin{equation}
\label{eq:int_params}
\hat{\bGamma} = [\hat{\bPhi}, \hat{\bTheta}, \hat{\bU}] = \mathcal{H}([\bPhi,\bTheta, \bU]) = \mathcal{H}(\bGamma),
\end{equation}
In other words, we first recover the latent variables using $\mathcal{F}$ and then apply a suitable amount of smoothing to them to obtain $\hat{\bGamma}$.
Finally, we apply $\mathcal{G}$ on $\hat{\bGamma}$ to obtain a new trajectory which then serves as a soft constraint during the next iteration of bundle adjustment.

In our experiments, we expect the UAV to move slowly and smoothly. Therefore in our current implementation, we used $\mathcal{H}(\bGamma) = (g \ast \bGamma)$, where $g$ denotes a Gaussian kernel. Other more sophisticated forms of $\mathcal{H}(\cdot)$ are worth exploring in the future. We can now write down the expression for the dynamics-based prior or regularizer.
\begin{equation}
\label{eq:prior}
R(\bX,\bGamma) =
\begin{bmatrix}
1 \\ \lambda_1 \\ \lambda_2 \\ \lambda_3 \\
\end{bmatrix}^\intercal
\begin{bmatrix}
(\bX - \mathcal{F}(\bGamma))^2 \\
\rho_1(\bPhi - \hat{\bPhi}) \\
\rho_2(\bTheta - \hat{\bTheta}) \\
\rho_3(\bU - \hat{\bU}) \\
\end{bmatrix}
\end{equation}
\noindent

%% file: optimization3.tex

\section{Optimization}
\label{sec:opt}

We now describe the steps needed to solve the regularized bundle adjustment problem stated in \eqs{optimization_problem}{prior}.
This is done using an efficient nonlinear least squares solver~\cite{ceres-solver}
and suitable initialization. In conventional SfM, the 3D points are treated independently resulting in a sparse problem that can be solved using a sparse Levenberg-Marquardt (LM) method~\cite{Marquardt1963}. 
As we discussed, imposing our dynamics-based regularizer directly would lead to a dense linear systems within each iteration of LM~\cite{Hartley00}.

\begin{algorithm}[t!]
	\begin{algorithmic}[1]
		\STATE \hl{Inputs}:
		\begin{itemize}[noitemsep,nolistsep]
			\item Initial trajectory $\bX^0$ and camera parameters $\bC^0$
			\item Observations in camera views $\bP$
		\end{itemize}
		\STATE \hl{Outputs}: Final estimates $(\bX^{*},\bC^{*},\bGamma^{*})$ and $(\bU_\theta,\bU_\phi)$
		\vspace{0.2cm}
		\FOR {iteration $s \in [1..S]$}
		\STATE $\bGamma^{s-1} \leftarrow \mathcal{G}(\bX^{s-1})$
		\STATE $\hat{\bGamma}^{s-1} \leftarrow \mathcal{H}(\bGamma^{s-1})$, $h(\cdot)$ defined in \eqt{int_params}
		\vspace{0.1cm}
		\STATE $(\bX^{s}, \bC^{s},\bGamma^{s}) \leftarrow$ run one step of LM to solve \eqt{optimization_problem}
		\ENDFOR
		\STATE $(\bX^{*},\bC^{*},\bGamma^{*}) \leftarrow (\bX^{s}, \bC^{s},\bGamma^{s})$
		\STATE $(\bU_\theta,\bU_\phi) \leftarrow (\bTheta,\bPhi)$ from \eqs{body_rates}{user_commands}
	\end{algorithmic}
	\caption{Bundle Adjustment with motion dynamics}
	\label{alg:optimization}
\end{algorithm}

Here, we describe our proposed technique to impose the regularization indirectly. We will use the trajectory estimate from the previous BA iteration to generate the soft constraints for the dynamics-based prior. Formally these constraints are represented by $\hat{\bGamma}^{s-1} = \mathcal{H}(\bGamma^{s-1})$, therefore we rewrite \eqt{prior} as:
\begin{equation}
\label{eq:prior_our}
R(\bX^s,\bGamma^s) =
\begin{bmatrix}
1 \\ \lambda_1 \\ \lambda_2 \\ \lambda_3 \\
\end{bmatrix}^\intercal
\begin{bmatrix}
(\bX^s - \mathcal{F}(\bGamma^s))^2 \\
\rho_1(\bPhi^s - \hat{\bPhi}^{s-1}) \\
\rho_2(\bTheta^s - \hat{\bTheta}^{s-1}) \\
\rho_3(\bU^s - \hat{\bU}^{s-1}) \\
\end{bmatrix}
\end{equation}
\noindent
where $s$ is the iteration index in the LM technique for solving \eqt{optimization_problem}, $\lambda_i \in \Real, i \in \{1,2,3\}$ are scalar weights and $\rho_i(\cdot) : \Real^2 \rightarrow \Real, i \in \{1,2,3\}$ are robust cost functions.
This allows us to make the points $\bx^s_t$ independent of each other in the current iteration. However, there is an indirect dependence on the associated points from the previous iteration $\bx^{s-1}_t$. \alg{optimization} depicts the exact steps of our method. So far, we have not discussed initialization of the camera poses and parameters. This is described next.

\vspace{0.1cm} \noindent \textbf{Trajectory and Camera Pose Initialization.}
First, we estimate camera poses $\bC^0$ using a traditional point-based SfM pipeline~\cite{Agarwal10}. Because the cameras are often far apart and the visible backgrounds do not overlap substantially, we have
used an additional camera to recover the initial calibration. We capture a video walking around the scene using an additional hand-held camera. Keyframes were extracted from this video and added to the frames selected from the fixed ground cameras. After running SfM on these frames, we extract the poses for the fixed cameras.
\begin{algorithm}[t!]
	\begin{algorithmic}[1]
		\STATE \hl{Inputs } :
		\begin{itemize}[noitemsep,nolistsep]
			\item Cameras $\bC : \{\bc_j\}, j \in [1.. M]$
			\item Sets of observations $\bp_{jt}: \{\bp_{jtk}\}, k \in [1..K_{jt}]$ for camera $j$ at time $t$
		\end{itemize}
		\STATE \hl{Outputs}: 3D Point $\bx_t$
		\vspace{0.3cm}
		\FOR {$i \in [1..N]$}
		\STATE Randomly pick $2$ cameras: $\bc_m, \bc_n$
		\STATE Randomly pick observation $\bp_{mtk}$ from $\bp_{mt}$
		\STATE Randomly pick observation $\bp_{ntl}$ from $\bp_{nt}$
		\STATE $x_i \leftarrow \mycode{triangulate-2view}(\bp_{mtk}, \bc_m, \bp_{ntl}, \bc_n)$
		\STATE $e(x_i) \leftarrow$  evaluate score for hypothesis $x_i$ using \eqt{repro_cost_min}
		\ENDFOR
		\STATE $\bx_t = \mathop{\arg\min} \limits_{x_i}(e(x_i))$
	\end{algorithmic}
	\caption{RANSAC-based multi-view triangulation}
	\label{alg:triang}
\end{algorithm}

Given $\bC^0$, we triangulate the trajectory points from detections obtained from background subtraction. We propose a robust RANSAC-based triangulation method to obtain $\bX^0$ (see \alg{triang}). This involves randomly picking a camera pair and triangulating two random detections in these cameras using a fast triangulation routine~\cite{Lindstrom10} to obtain a 3D point hypothesis.
Amongst all the hypotheses, we select the 3D point that has the lowest total residual error in all the views (as defined in Eq.~\ref{eq:repro_cost_min}. We use \alg{triang} on all timesteps to compute the initial trajectory $\bX^0$.

\vspace{0.1cm} \noindent \textbf{Implementation details.}
Our method is implemented in $\mycode{C++}$, using the Ceres non linear least squares minimizer~\cite{ceres-solver}. In order to detect the UAV we have used the OpenCV~\cite{OpenCV} implementation of Gaussian Mixture Models (GMM)-based background subtraction~\cite{KaewTraKulPong2002}. Briefly, it creates a GMM model for the background and updates it with each incoming frame. Further, the regions of the image that are not consistent with the model are considered to be foreground and we use them as detections. This, however, leads to large amounts of false positives for the outdoor videos, therefore we processed the resulting detections with a Kalman Filter (KF)~\cite{Welch90} with a constant acceleration model. Detections from the resulting tracks are used in our BA optimization. To reduce the amount of false positives we only considered tracks that are longer than 3 time steps. As a result we ended up with 0-8 detections per frame in all the camera views (see \fig{real_outdoor_exp}).


%% file: experiments.tex

%
%

\begin{figure}[t!]
	\centering
	\includegraphics[width = 0.95\linewidth]{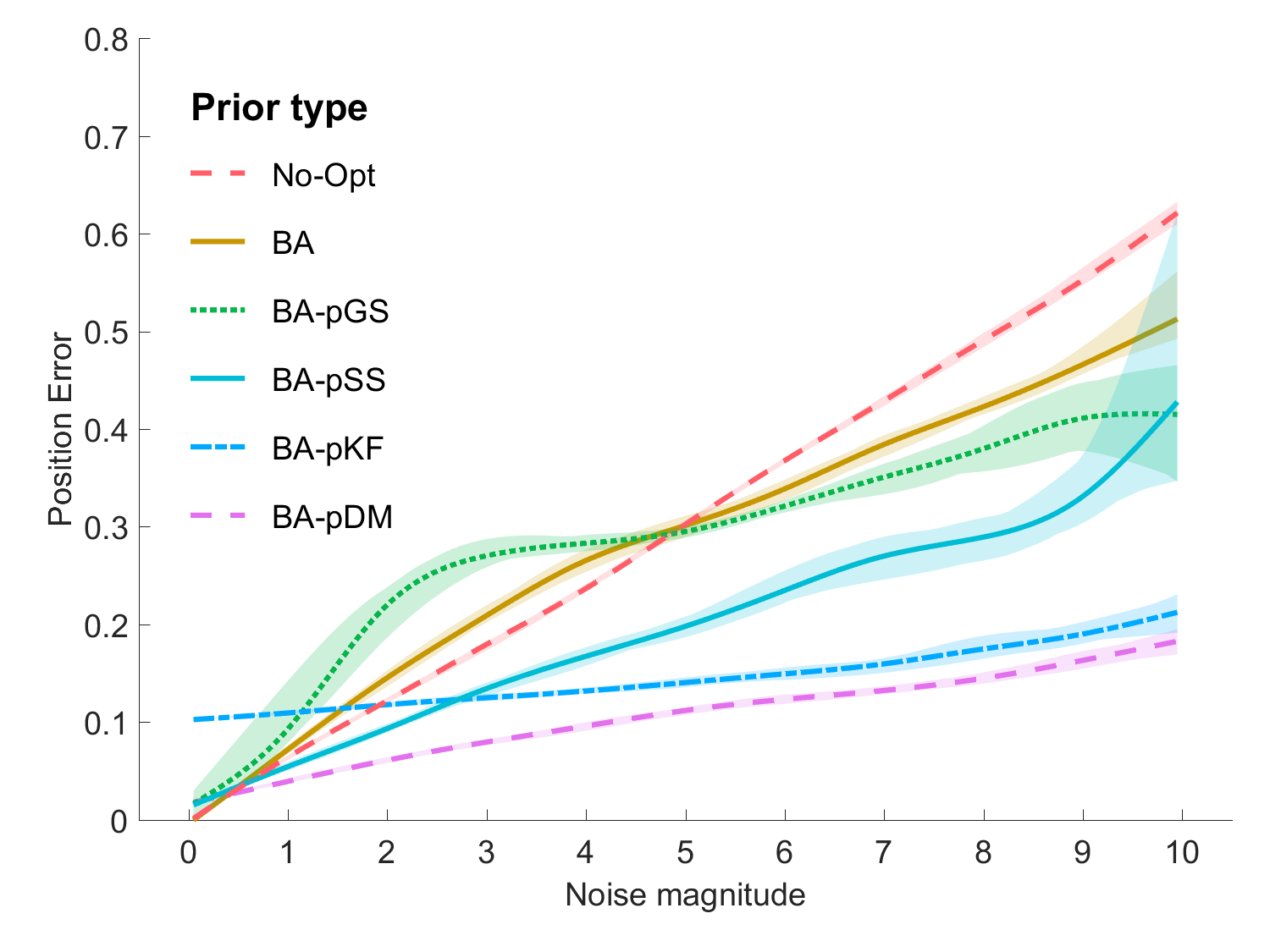}\\
	\vspace{-0.3cm}
	\caption{[Evaluation on synthetic data] with respect to different amount of noise added to the initial point locations, camera parameters and point observations in camera views. For each method plot above shows mean and standard deviation of the resulting trajectory from the ground truth across 10 different runs of our algorithm. (best seen in color)}
	\label{fig:synth_eval}
	\vspace{-0.4cm}
\end{figure}

\begin{figure*}[t!]
	\centering
	\includegraphics[width = 0.9\linewidth]{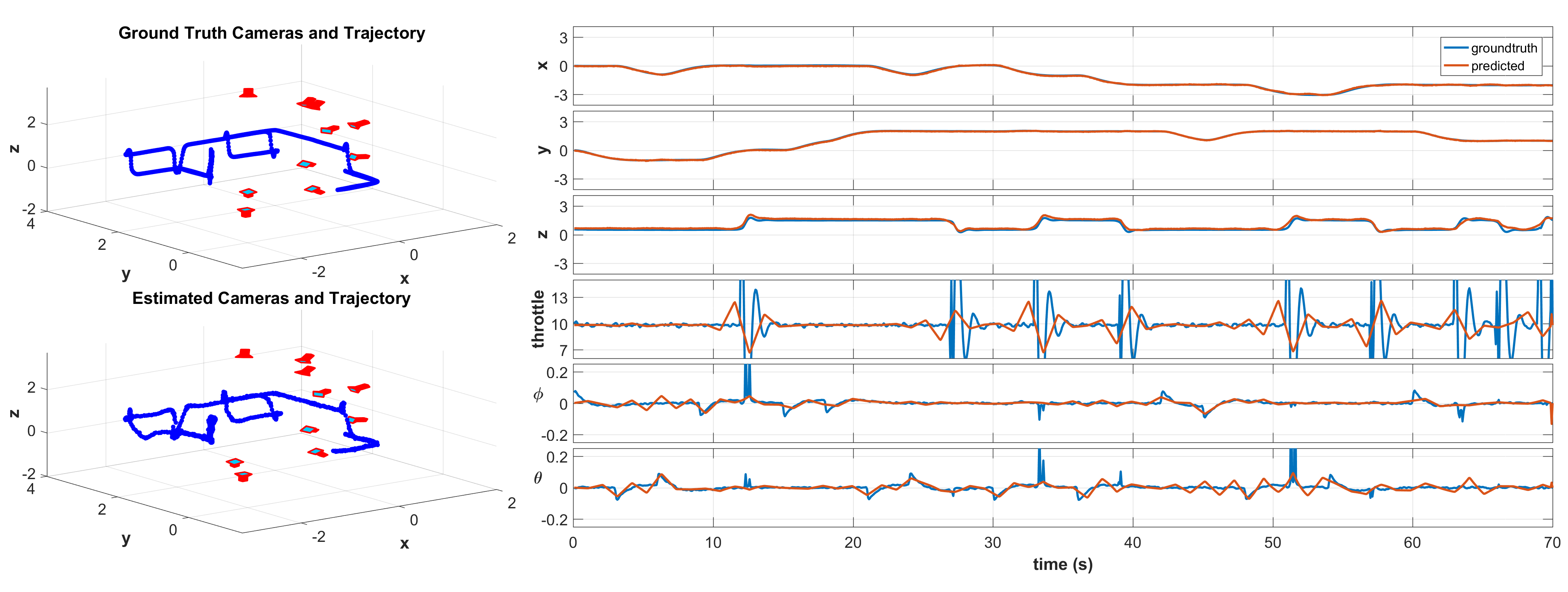}
	\vspace{-0.5cm}
	\caption{Comparison between the predicted and ground truth positions and internal parameters of the quadrotor. In the left part of the figure you can see the visualization of trajectory point locations. On the right we can see the difference between 3D coordinates $(\bX)$ of the quadrotor and its internal parameters $(\bU, \bPhi,\bTheta)$. (best seen in color)}
	\label{fig:synth_dyn_eval}
	\vspace{-0.5cm}
\end{figure*}

\section{Experiments}
\label{sec:results}
We first evaluated our proposed method on synthetic data obtained from a
realistic quadrotor flight simulator to analyze accuracy and robustness of our
estimated trajectories and control inputs in the presence of image noise,
outliers in tracking and errors in the initial camera pose parameters.
We also present several results on real data captured indoors and a large outdoor
scene where we have access to ground truth trajectory information.

We measure the accuracy of our estimated trajectories by robustly aligning our trajectory 
estimate to the ground truth trajectory~\cite{Horn87}. In the ground truth coordinate
system, we then calculate the root mean squared error (RMSE). We have compared many variants
of BA. The suffix `-p' below denotes the type of regularizer (prior).
\vspace{-0.10cm}
\begin{itemize}[leftmargin=*]
	\setlength{\itemsep}{0pt}
	\setlength{\parskip}{2pt}
	\item \hl{No-Opt}: has no bundle adjustment optimization.
	\item \hl{BA}: does regular point-based BA without regularization.
	\item \hl{BA-pGS}: uses a Gaussian smoothing based regularizer.
	\item \hl{BA-pSS}: uses a spline smoothing based regularizer.
	\item \hl{BA-pKF}: This denotes the method with a Kalman filter based motion regularizer (see Appendix~\ref{app:kalman}).
	\item \hl{BA-pDM}: This denotes our proposed method with the flight dynamics-based regularizer.
\end{itemize}

\noindent \textbf{Datasets.}
We evaluated our method on three sets of data. We used an existing quadrotor simulator\footnote{\url{ github.com/OMARI1988/Quadrotor_MATLAB_simulation}} to generate trajectories with random waypoints. Each simulated trajectory contains $510$ points or time steps that was used to simulate 17 seconds of video at 30Hz from 10 cameras. The camera locations were randomly generated around the flight volume and the cameras were oriented towards the center of the flight volume. We generated $100$ sequences with varying degree of noise in the camera pose, initial trajectories as well as image noise and outliers.

\begin{figure}[t]
	\centering
	\begin{tabular}{cc}
		\hspace{-0.25cm} 
		\includegraphics[height=1.8cm,width=0.485\linewidth]{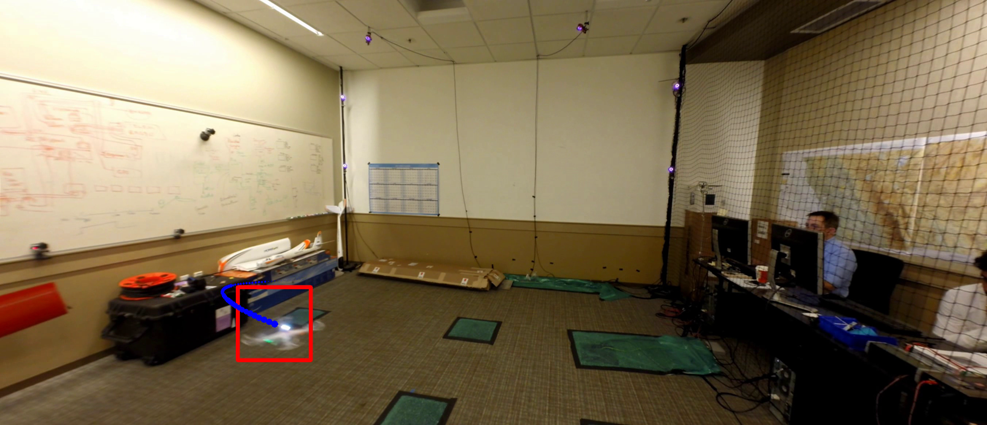} &
		\hspace{-0.25cm} 
		\includegraphics[height=1.8cm,width=0.485\linewidth]{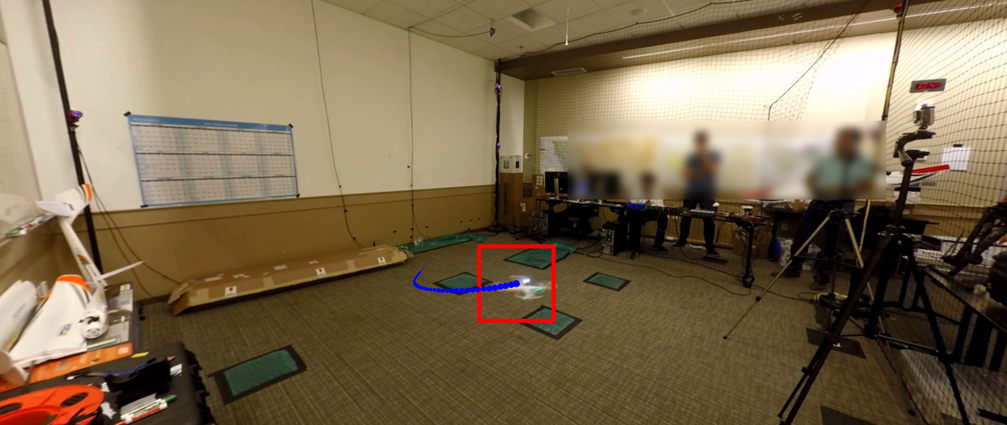} \\
		\multicolumn{2}{c}{\begin{tabular}{cccccc}
				\hspace{-0.17cm}\includegraphics[width=0.10\linewidth]{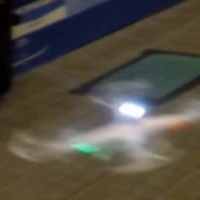} &
				\hspace{-0.25cm}\includegraphics[width=0.10\linewidth]{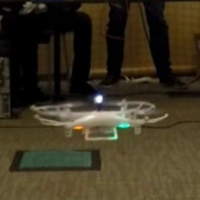} &
				\hspace{-0.25cm}\includegraphics[width=0.10\linewidth]{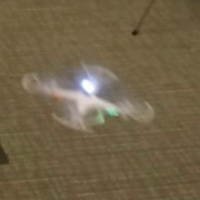} &
				\hspace{-0.25cm}\includegraphics[width=0.10\linewidth]{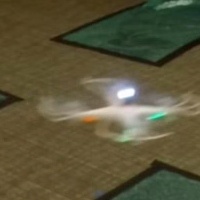} &
				\hspace{-0.25cm}\includegraphics[width=0.10\linewidth]{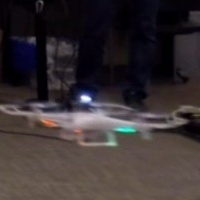} &
				\hspace{-0.25cm}\includegraphics[width=0.10\linewidth]{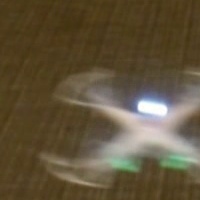} \\
				\hspace{-0.25cm}\includegraphics[width=0.10\linewidth]{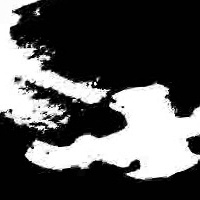} &
				\hspace{-0.25cm}\includegraphics[width=0.10\linewidth]{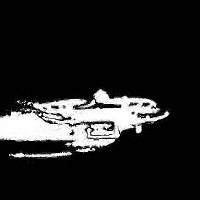} &
				\hspace{-0.25cm}\includegraphics[width=0.10\linewidth]{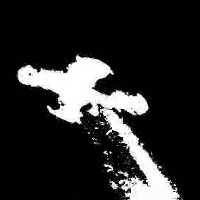} &
				\hspace{-0.25cm}\includegraphics[width=0.10\linewidth]{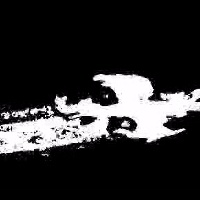} &
				\hspace{-0.25cm}\includegraphics[width=0.10\linewidth]{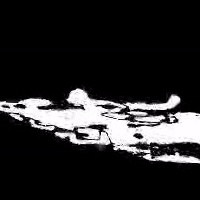} &
				\hspace{-0.25cm}\includegraphics[width=0.10\linewidth]{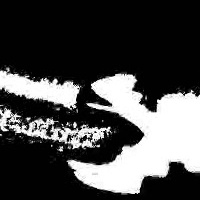} \\
			\end{tabular}
		} \\
		\multicolumn{2}{c}{
			\includegraphics[width=0.85\linewidth]{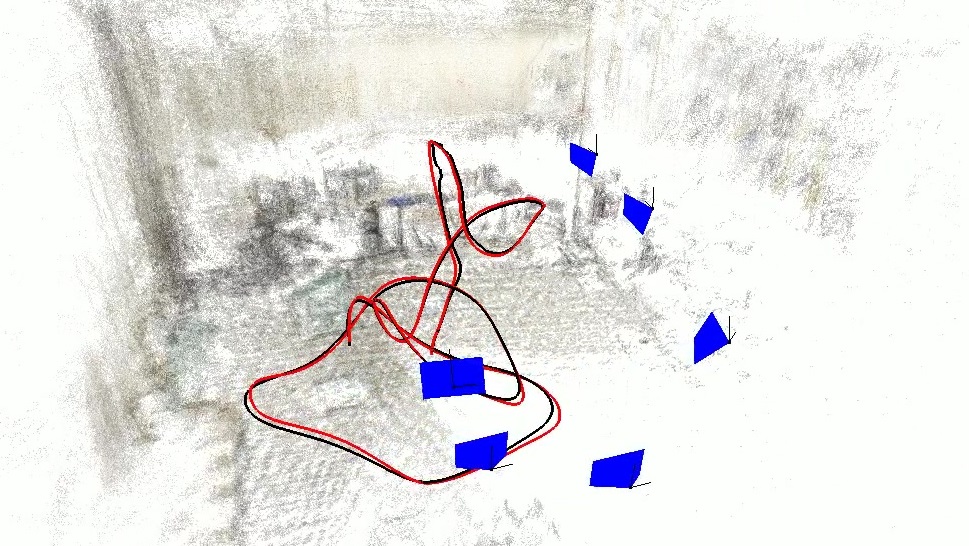} 			
		} \\
	\end{tabular}
	\caption{Qualitative results (\textsc{Lab} dataset): [\textsc{Top}] Two of the six camera viewpoints. [\textsc{Middle}] Zoomed in patches of the trajectory point from all camera views with corresponding background segmentation results. [\textsc{Bottom}] A 3D visualization of the estimated trajectory (in black) and the ground truth trajectory (in red). (best seen in color)}
	\label{fig:real_exp}
	\vspace{-0.6cm}
\end{figure}
\begin{figure*}[ht]
	\centering
	\begin{tabular}{cc}
	\includegraphics[width = 0.73\linewidth]{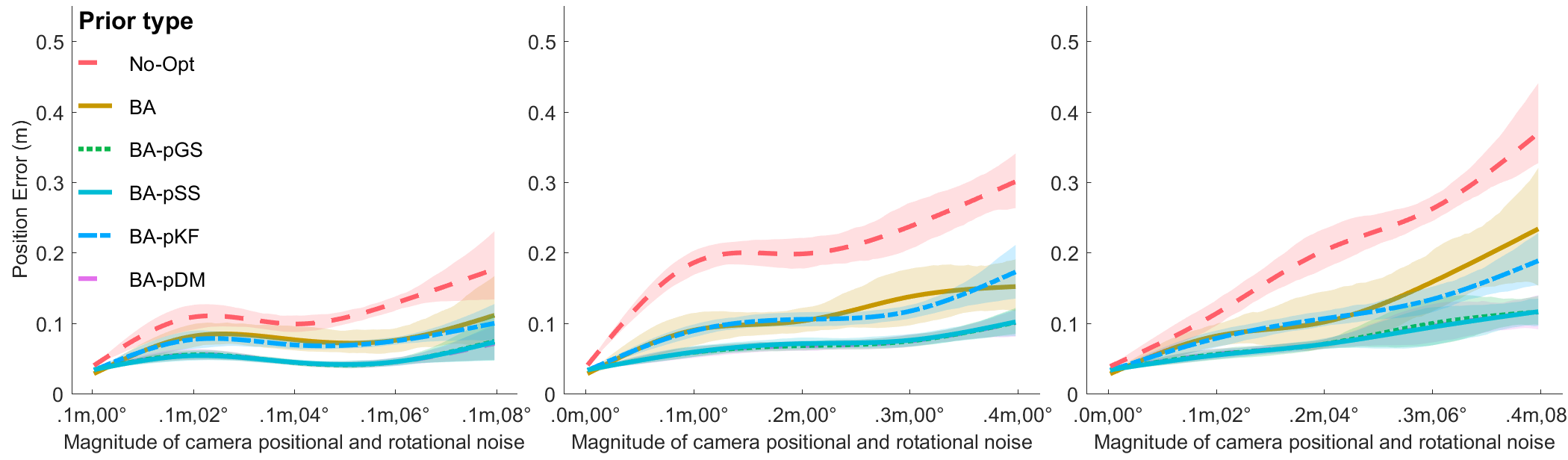} &
	\raisebox{1.6cm}{\begin{tabular}{lc}
			\toprule
			\scriptsize{method} & \scriptsize{Position Error(m)} \\
		\hl{No-Opt} & .2045$\pm$.015 \\
		\hl{BA}     & .1165$\pm$.014 \\
		\hl{BA-pGS} & .0760$\pm$.007 \\
		\hl{BA-pSS} & .0761$\pm$.007 \\
		\hl{BA-pKF} & .1141$\pm$.009 \\
		\hl{BA-pDM} & \textbf{.0757$\pm$.007} \\
		\bottomrule
	\end{tabular}} \\
	\end{tabular}
	\caption{ Comparing the various methods on  the \textsc{Lab} dataset. The initial camera poses are perturbed with increasing position and orientation noise. The plots show the mean and std. dev. of the final error from $10$ runs for the various methods. The table lists the error statistics (mean and std. dev.) for all the methods across all the different runs. (best seen in color)}
	\label{fig:real_noise_eval}
	\vspace{-0.5cm}
\end{figure*}
\begin{figure}[t]
	\centering
	\includegraphics[width = \linewidth]{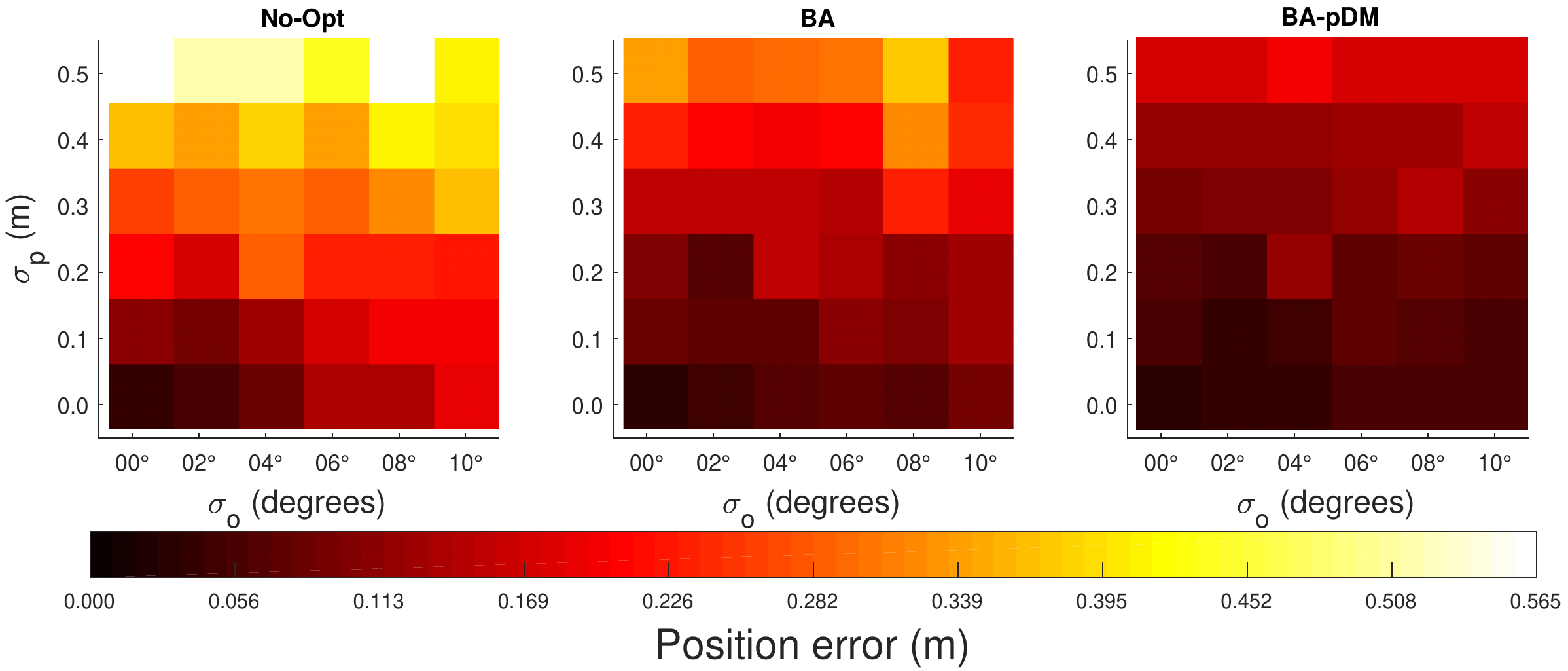}
	\vspace{-0.6cm}
	\caption{Sensitivity analysis on \textsc{Lab} dataset.  [\textsc{Left}] Triangulation result. [\textsc{Middle}] Results of standard BA~\cite{Agarwal10} without any prior information. [\textsc{Right}] Results of our method with a dynamics-based prior. Each of these plots show the mean position error across $10$ runs, for different amounts of noise added to the camera positions ($\sigma_p$) and orientations ($\sigma_o$). The amount of error is color-coded and the colorbar below the plots show the correspondence between colors and actual values in meters. (best seen in color)}
	\label{fig:real_heatmap}
	\vspace{-0.5cm}
\end{figure}

We evaluated our method on two datasets captured in real scenes -- \textsc{Lab} and  \textsc{Farm}. In both cases, six GoPro tripod-mounted cameras were pointed in the direction of the flight volume. We  recorded video at $2704 \times 1536$ pixel resolution and 30 fps. The \textsc{Lab-Scene} was approximately $6 \times 8$ meters and contains an OptiTrack motion capture system~\cite{Optitrack} that was used to collect the ground truth quadrotor trajectory and camera poses. In this case we flew an off-the-shelf quadrotor (44.5 cm diameter) for 35 seconds. In order to achieve precise tracking accuracy we equipped UAV with a bright LED and an OptiTrack marker, placed close to each other. LED helps us to facilitate the detection process, as now it is narrowed down to searching for the brightest point in the frame. Despite the simple scenario, people that were moving around in the room and reflectance from the walls were frequently detected as moving objects.

The \textsc{Farm} dataset was captured in a large outdoor scene (see Figure~\ref{fig:teaser}) with a bigger UAV (1.5 m in diameter). Footage of a 4-minute flight was recorded using the six cameras, placed $\sim20$ meters apart. The videos were captured at the frame-rate of $15$ FPS, which results in $3600$ trajectory points. We have also collected drone GPS data and used it as ground truth for evaluation. 

\subsection{Experiments on Synthetic Data}
\label{sec:exp_synth}
Here we first describe the influence of noise on the performance of our method. We then show that our approach is capable of inferring the underlying control inputs that define the motion of the drone.

\vspace{0.1cm}
\noindent \textbf{Sensitivity Analysis.}
\fig{synth_eval} summarizes the results on the synthetic dataset. We see that prior information mostly helps to improve reconstruction accuracy. Of all the priors, motion-based ones show a significant improvement over conventional smoothing methods. Overall, the dynamics-based prior is the most accurate. This is probably because the trajectory generated by the quadrotor simulator complies with the same model used to develop the dynamics-based prior (\sect{quad_dynamics}).

\vspace{0.1cm}
\noindent \textbf{Inferring Control Inputs.}
We have evaluated the quality of prediction of the internal parameters $(\bU, \bPhi,\bTheta)$ of the quadrotor on the synthetic dataset, as it provides an easy way of collecting ground truth information for these parameters. \fig{synth_dyn_eval} summarizes this comparison. In the left part of the figure we can see the predicted and ground truth locations of the 3D trajectory point locations. The right part of the figure depicts the difference between predicted and ground truth $(x,y,z)$ positions of the quadrotor in the environment and the comparison of the predicted internal parameters of the quadrotor with the ground truth.

\fig{synth_dyn_eval} shows that 3D locations of the quadrotor were predicted quite well with very small deviations from the ground truth. Regarding the internal parameters, our method is able to predict them very well for the parts of the sequence, where these parameters vary smoothly. This behavior is introduced by \alg{optimization}, where we constrain the values of $(\bU,\bPhi,\bTheta)$ to vary smoothly through time. \comment{We provide a more thorough evaluation of influence of the prior parameters on the inference of $(\bU,\bPhi,\bTheta)$ in the Appendix~\ref{app:control}. }In our future research we would like to investigate other constraints on the internal quadrotor parameters that will allow us to recover their sharp changes.

\subsection{Results on \textsc{Lab} Dataset}

\begin{figure*}[t!]
	\centering
	\begin{tabular}{cc}
		\includegraphics[width = 0.42\linewidth]{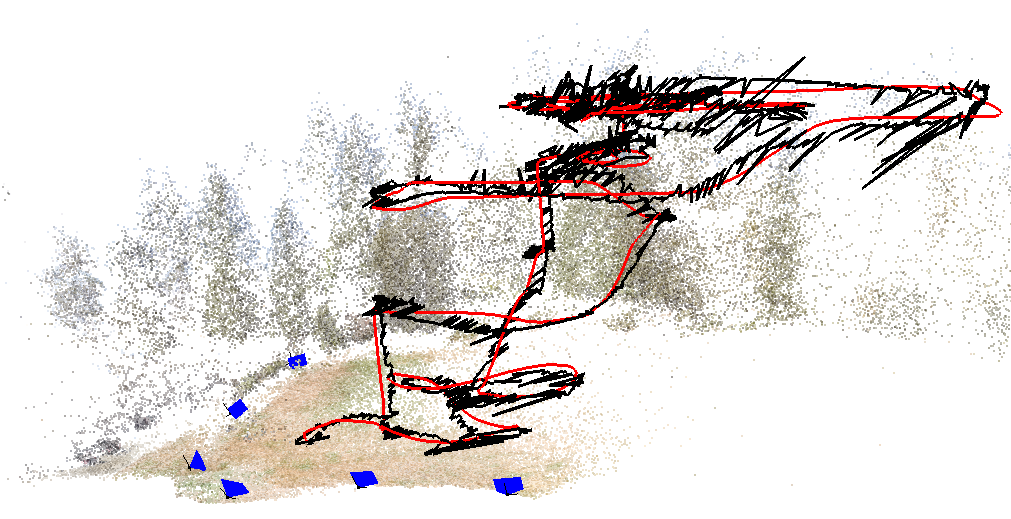} &
		\includegraphics[width = 0.42\linewidth]{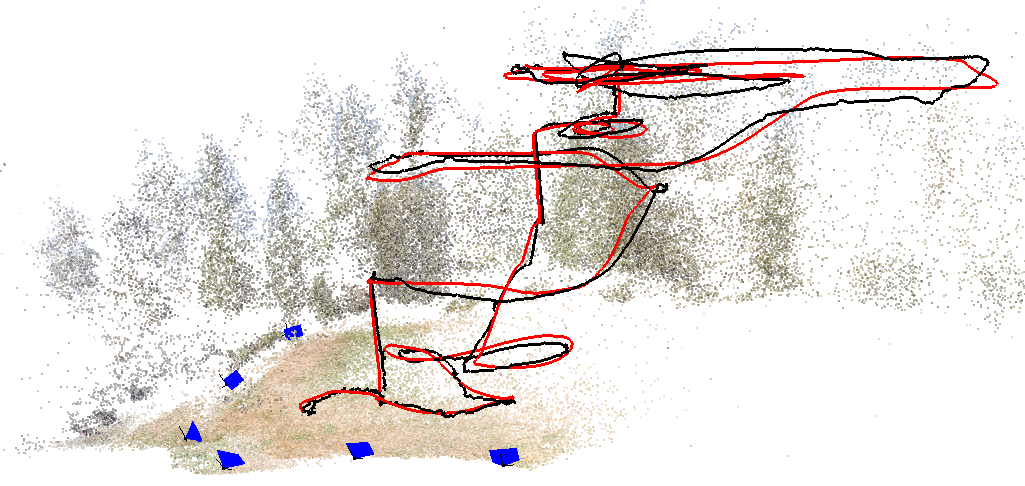} \vspace{-0.2cm} \\
		\multicolumn{2}{c}{\hspace{-0.3cm}\includegraphics[width = \linewidth]{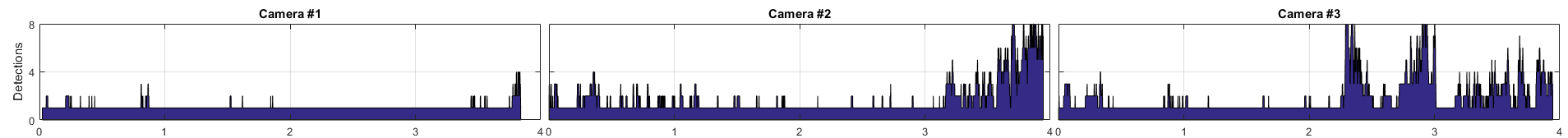}} \\
		\vspace{-0.7cm}	
	\end{tabular}
	\caption{Qualitative Results (\textsc{Farm} dataset): The \hl{black} and \hl{red} curves denote the estimated and ground truth (GPS) trajectory respectively. [\textsc{Left}] Initialization trajectory estimate after the triangulation step. [\textsc{Right}] The result obtained after bundle adjustment (BA-pDM). The trajectory is smoother and more accurate compared to the initial trajectory. [\textsc{Bottom}] The number of per-frame candidate detections for 3 videos. The middle and right plots show where tracking was difficult.}
	\label{fig:real_outdoor_exp}
	\vspace{-0.5cm}	
\end{figure*}
\fig{real_exp} depicts an example of our indoor experiment. \fig{real_exp}(top) depicts sample camera views. \fig{real_exp}(middle) illustrates the cropped and zoomed in patches around the tracked object. \fig{real_exp}(bottom) shows the 3D reconstruction of the trajectory compared to the OptiTrack ground truth.

We also performed a quantitative evaluation of our method. In the \textsc{Lab} dataset, the detections in individual frames were quite reliable. Therefore to perform the noise sensitivity analysis, we have added noise to our initial camera pose estimates before running the optimization.

\figs{real_noise_eval}{real_heatmap} show the quantitative evaluation results. Note that as we increase the noise added to the initial camera poses, the triangulation accuracy steadily decreases. This is because the point correspondence between different camera views progressively become inaccurate. The use of standard bundle adjustment improves the reconstruction accuracy. However, the quadrotor dynamics-based prior produces even higher accuracy especially when the noise is quite significant (see \fig{real_heatmap}).

In \fig{real_noise_eval}, the plots for BA-pDM, BA-pGS and BA-pSS methods are almost indistinguishable from each other, due to very similar performance. This is because in the indoor experiments, the UAV is always clearly visible and not too far from all the cameras. This results in high quality detections with few false positives and allows even BA with simple smoothing-based priors to achieve good accuracy.
	
The general trend we noticed in these experiments
was that when the initial camera parameters are relatively accurate
even simple triangulation can produce quite reliable trajectories. However,
larger errors in initial camera pose quickly degrades the performance of
both standard triangulation and bundle adjustment methods,
whereas our approach is robust to relatively larger amounts of
camera pose error due to the effective use of quadrotor dynamics.
\subsection{Results on \textsc{Farm} Dataset}


Finally we have evaluated our methods on the outdoor dataset. \fig{teaser} depicts an example of our outdoor experiment. We can see that compared to the indoor case the drone is much further away from the camera, which results in some challenges not just for the optimization, but also for its detection with background subtraction algorithms. For this experiment we have used \alg{triang} to set the initial values for the camera parameters and trajectory point locations. \fig{real_outdoor_exp}(left) shows the result obtained using \alg{triang}, which we use as an initialization. \fig{real_outdoor_exp}(right) shows the final result of our approach. \fig{real_outdoor_exp}(bottom) depicts the number of detections per frame for three out of six camera views. We can see that towards the end of the sequence more false positives appear, due to the UAV entering a complicated area of the environment. Nevertheless, our approach allows successfully tracking the drone. \comment{Some additional examples of our method dealing with complicated backgrounds can be found in the supplementary material.}
\begin{figure}[t]
	\centering
	\begin{tabular}{c}
		\includegraphics[width = 0.9\linewidth]{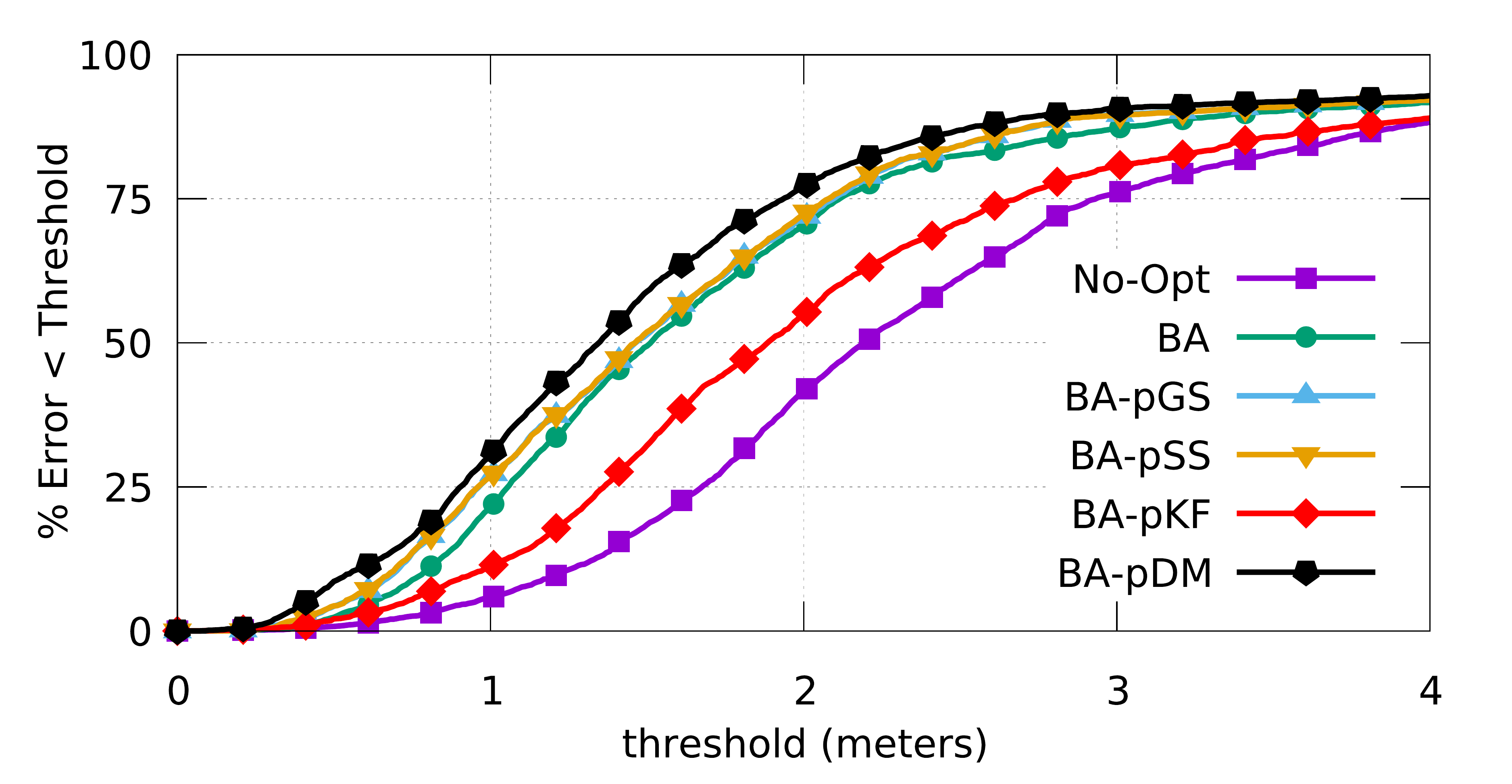} \\
		\hspace{-0.3cm}
		\begin{tabular}{cccccc}
			\toprule
			\hspace{-0.3cm} No-Opt \hspace{-0.1cm}
			& \hspace{-0.2cm} BA \hspace{-0.1cm}
			& \hspace{-0.2cm} BA-pGS \hspace{-0.1cm}
			& \hspace{-0.2cm} BA-pSS \hspace{-0.1cm}
			& \hspace{-0.2cm} BA-pKF \hspace{-0.1cm}
			& \hspace{-0.2cm} BA-pDM \hspace{-0.3cm}
			\\
			\cmidrule{1-6}
			2.551 & 1.910 & 1.785 & 1.781 & 2.275 & {\bf 1.636} \\
			\bottomrule
		\end{tabular}
		\\
	\end{tabular}
	\caption{Comparing various methods on the \textsc{Farm} dataset. The percentage of 3D points with position error less than a threshold is shown for a range of thresholds. The RMSE error in meters for all the methods are reported in the table.}
	\label{fig:outdoor_results}
	\vspace{-0.5cm}
\end{figure}

\fig{outdoor_results} summarizes the accuracy of the various methods. Similar to the other experiments, the prior information helps to recover a more accurate trajectory. Moreover, using an appropriate dynamics model prior produces the most accurate result amongst all the priors.

The Kalman filter-based prior is not very effective in this case due to noise in the initial trajectory (see \fig{real_outdoor_exp}(left)). The initialization noise causes the Kalman filter to be unstable and it fails to recover the correct motion model parameters unlike the other priors which are more robust.


%% file: appendix.tex

\section*{Supplementary Appendix}
\appendix

In this section we present more details on one of the baseline methods ({\bf BA-pKF}). We also report two additional experiments. In the first one, we show a sensitivity analysis regarding parameter settings in our proposed {\bf BA-pDM} method and discuss how this affects the accuracy of the estimated control inputs. In the second experiment, we compare our BA procedure with a baseline where the data association problem was solved before running bundle adjustment by forcing the single-view tracker to output at most one 2D observation in every video frame. \comment{Finally, qualitative results on the on the \textsc{Lab} and  \textsc{Farm} datasets are shown in the supplementary video.}

\section{Kalman filter prior}
\label{app:kalman}

In \sect{results}, we described several baselines that we have compared our approach to. Here we provide more details on the method based on the Kalman filter, as the other ones are relatively straightforward to implement.

Classical Kalman filter allows predicting the state of the quadrotor at time $t+1$ from its state at time $t$. In our case this state contains drone's position and velocity. We then track quadrotor in 3D using the constant acceleration Kalman filter model. Therefore, $\hat{\bx}_{t+1}$ from \eqt{prior_generic} is computed according to the prediction of the Kalman filter. However, as the experiments show, the prior is dependent on the quality of the initialization and \hl{BA-pKF} is not robust to noise in the initial 3D trajectory.

\section{Control inputs prediction analysis}
\label{app:control}

In \sect{exp_synth} we have shown that our system is capable of inferring the internal state of the quadrotor, which can be further used to estimate the control inputs commanded to the drone by the operator. \fig{synth_dyn_eval} shows that our approach tends to over smooth the internal parameters $(\bPhi,\bTheta,\bU)$. Therefore, in this section we investigated the effect of fine-tuning the dynamics-based prior on the accuracy of the estimated $(\bPhi,\bTheta,\bU)$.

\begin{figure*}
	\centering
	\begin{tabular}{c}
		\toprule
		\includegraphics[width = 0.95\linewidth]{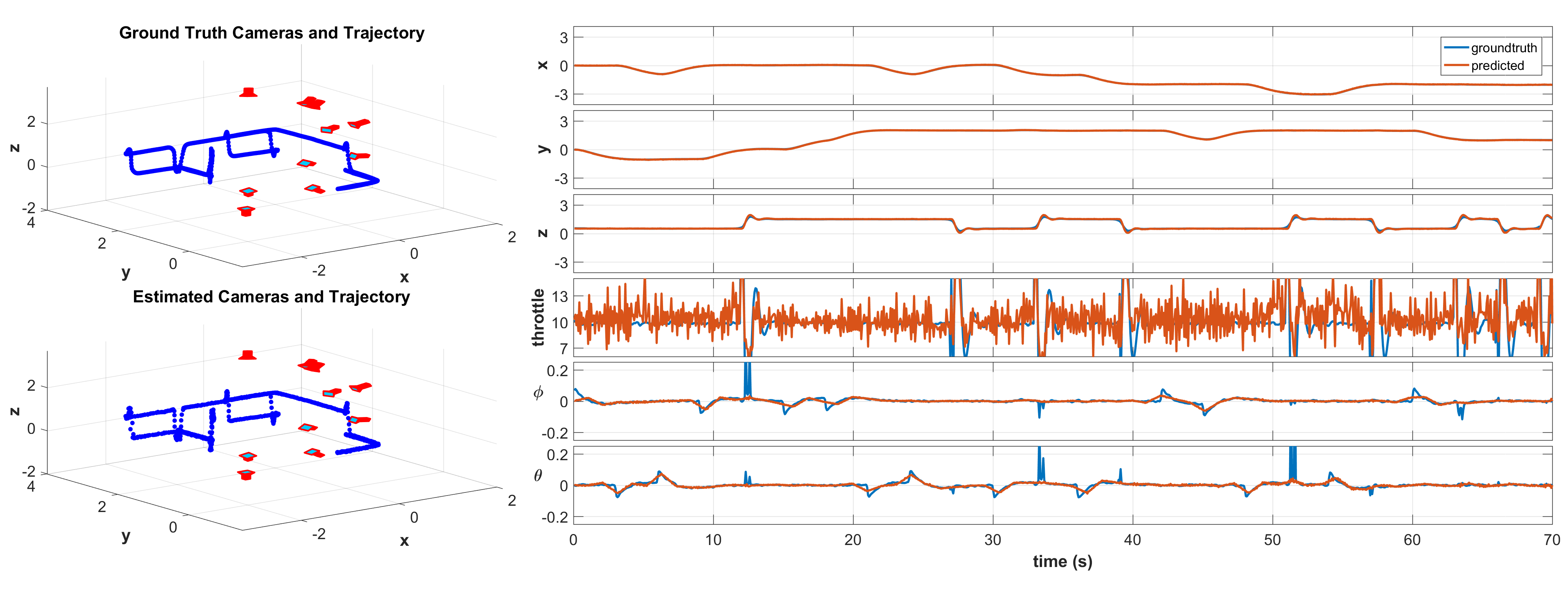} \\
		(a) $\lambda = 0.01, \sigma = 0.8$ \\
		\midrule
		\includegraphics[width = 0.95\linewidth]{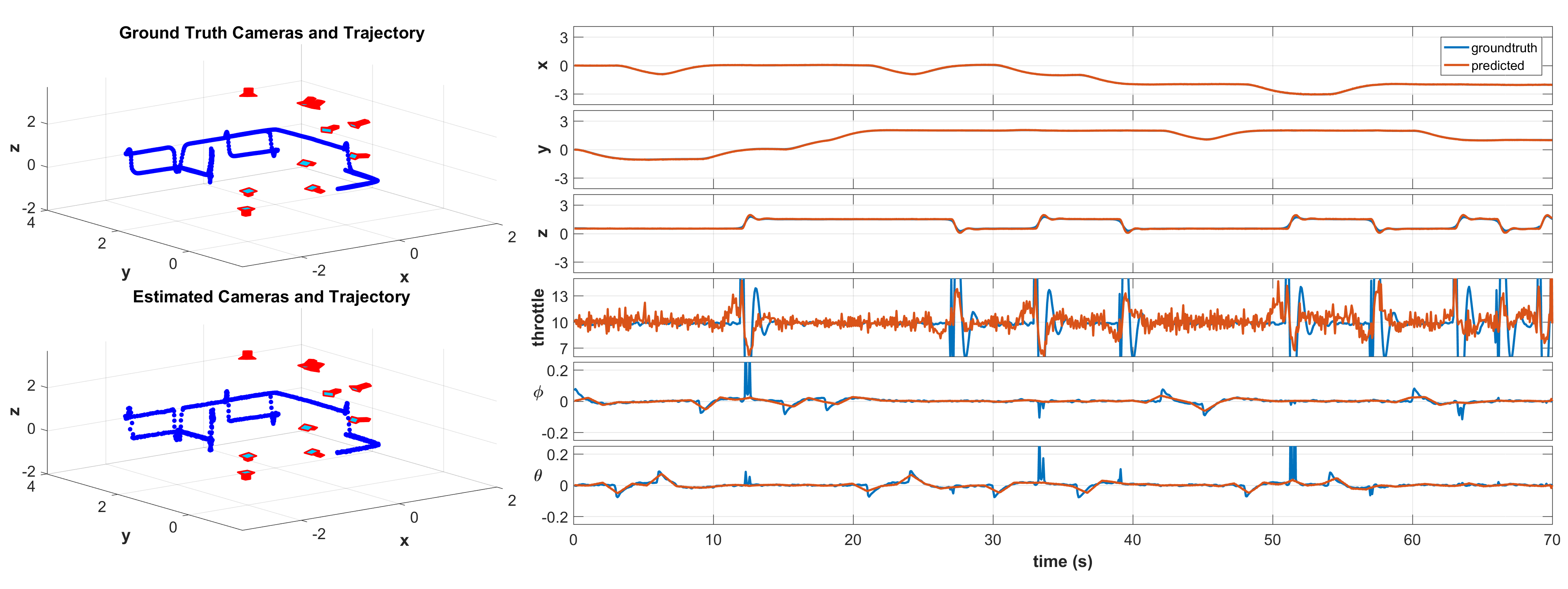} \\	
		(b) $\lambda = 0.02, \sigma = 1.1$\\
		\midrule
		\includegraphics[width = 0.95\linewidth]{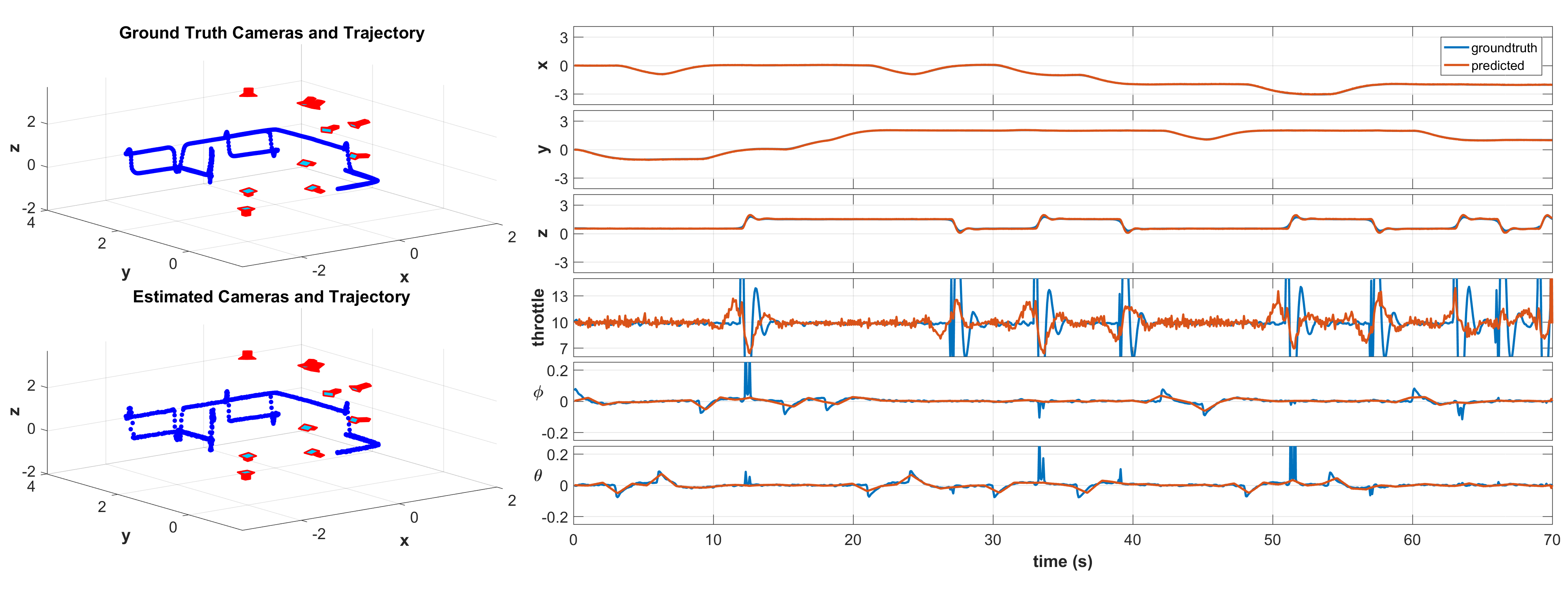} \\
		(c) $\lambda = 0.03, \sigma = 1.4$\\
		\bottomrule
	\end{tabular}
	\caption{Influence of the dynamics-based prior on the prediction of the internal state of the quadrotor.  In different plots we have varied the weight of the prior $\lambda$ and the smoothing factor $\sigma$ of the gaussian kernel that smooths $(\bPhi, \bTheta, \bU)$. (a) corresponds to $(\lambda,\sigma) = (0.01,0.8)$, (b) illustrates the case when $(\lambda,\sigma) = (0.02,1.1)$ and (c) depicts the experiment with $(\lambda,\sigma) = (0.03,1.4)$.}
	\label{fig:dynamic_prior_influence}
\end{figure*}

\fig{dynamic_prior_influence} illustrates the influence of the dynamic-based prior on internal quadrotor state. Here the weight $\lambda$ from the \eqt{optimization_problem} together with the smoothing coefficient $\sigma$ of the gaussian kernel $\mathcal{H}(\bGamma) = (g \ast \bGamma)$ from \eqt{int_params} increases from top most plot to the bottom one. We can see in \fig{dynamic_prior_influence}(a) that if the weight of the prior is small $(0.01)$, we can quite reliably recover the sharp peaks of the throttle ($\bU$) command, however, there is some residual noise for time periods when the true throttle command remains fixed at a constant value. This happens because the prior does not have enough influence on the optimization to make it robust to the measurement noise in the image. On the other hand increasing the weight of the prior to $(0.03)$ (\fig{dynamic_prior_influence}(c)) allows us to compensate for the initialization noise and recover a smoother estimation of $\bU$. However, this tends to oversmooth $\bU$ especially when it changes abruptly corresponding to times when the drone suddenly changes direction or altitude.
\begin{table}[t!]
	\centering
	\begin{tabularx}{\linewidth}{Xc}
		\toprule
		\scriptsize{method} & Position error (m) \\
		\cmidrule{2-2}
		BA-pDM-single       & $1.998$ \\
		BA-pDM 				& $1.636$ \\
		\bottomrule
	\end{tabularx}
	\caption{Comparison between our method (\hl{BA-pDM}) and a baseline (\hl{BA-pDM-single}) on the \textsc{Farm} dataset. The baseline uses at most one detection in every frame in all the input videos.}
	\label{tbl:comp2single_det}
\end{table}

\section{Advantage of the new cost function}
\label{app:cost}

Recall that in \eqt{repro_cost_min}, we introduced a cost function based on a more general form of reprojection error that did not rely on perfect correspondence between different views. We did this to handle multiple candidate detections in every frame produced by the single-view tracker running on the input videos. To quantify the benefit of this new cost function, we compared our method (\hl{BA-pDM}) which uses this
new cost function $E(C, X, O)$ with another baseline which we refer to as (\hl{BA-pDM-single}). This baseline uses at most one measurement (detection) in every video frame. In that case, $E(C,X,O)$ becomes identical to the cost function $E_{BA}(C, X, O)$ (see \eqt{repro_cost}) used in conventional bundle adjustment.
These unique detections used in the baseline, were selected during the 3D trajectory initialization step, which uses the RANSAC-based multi-view triangulation method we have proposed.

\tbl{comp2single_det} reports the final average position error for 3D points sampled on the trajectories estimated by our method (\hl{BA-pDM}) and by the baseline (\hl{BA-pDM-single}) respectively.
These correspond to the \textsc{Farm} dataset. Note that both methods use the same dynamics-based prior but \hl{BA-pDM} produces a more accurate result because
the new cost function allows the selection of the 2D measurements (amongst the multiple detection candidates) to be refined during the bundle adjustment procedure.